\documentclass[journal,twoside,web]{ieeecolor}
\usepackage{generic}
\usepackage{cite}
\usepackage{amsmath,amssymb,amsfonts}
\usepackage{graphicx}
\usepackage{textcomp}
\usepackage{booktabs}
\usepackage{multirow}
\usepackage{xcolor}
\usepackage{colortbl}
\usepackage{array}
\usepackage{tabularx}
\usepackage{url}\AtBeginDocument{\urlstyle{same}}
\providecommand{\refname}{References}
\usepackage{etoolbox}
\patchcmd{\thebibliography}{plus pt}{plus .5pt}{}{\typeout{TMI template patch not applied}}
\usepackage{hyperref}
\hypersetup{hidelinks=true}
\makeatletter
\let\GRIPincludegraphics\includegraphics
\renewcommand{\includegraphics}[2][]{\edef\GRIP@file{#2}\def\GRIP@logo{blank.eps}%
  \ifx\GRIP@file\GRIP@logo\hbox to 0.65242pt{\vrule width 0pt height 0.65242pt depth 0pt\hss}\else\GRIPincludegraphics[#1]{#2}\fi}
\makeatother
\def\logoname{blank}
\graphicspath{{sec/figures/}}

\begin{document}
\bstctlcite{IEEEexample:BSTcontrol}
\title{Privileged On-Policy Distillation for Calibrated Alzheimer's Diagnosis With Missing Modalities}
\author{Vo Minh Khoi Nguyen, Dat Tien Chung, Quang Minh Khoa Le, Duc Khang Nguyen, Le Thien Phuc Nguyen,
Hoang Thien Nguyen, Minh Huu Nhat Le, Phat K. Huynh, Cuong Tuan Nguyen, Thanh Huy Nguyen, and Ulas Bagci, \IEEEmembership{Senior Member, IEEE},
for the Alzheimer's Disease Neuroimaging Initiative\textsuperscript{*} and the Australian Imaging Biomarkers and Lifestyle
flagship study of ageing\textsuperscript{\dag}
\thanks{\textsuperscript{*}Data used in preparation of this article were obtained from the Alzheimer's Disease
Neuroimaging Initiative (ADNI) database (adni.loni.usc.edu). As such, the investigators within the ADNI contributed
to the design and implementation of ADNI and/or provided data but did not participate in analysis or writing of this
report. A complete listing of ADNI investigators can be found at:
\protect\url{http://adni.loni.usc.edu/wp-content/uploads/how_to_apply/ADNI_Acknowledgement_List.pdf}.}
\thanks{\textsuperscript{\dag}Data used in the preparation of this article was obtained from the Australian Imaging
Biomarkers and Lifestyle flagship study of ageing (AIBL) funded by the Commonwealth Scientific and Industrial Research
Organisation (CSIRO) which was made available at the ADNI database (www.loni.usc.edu/ADNI). The AIBL researchers
contributed data but did not participate in analysis or writing of this report. AIBL researchers are listed at
data.aibl.org.au/adni/.}
\thanks{V.~M.~K. Nguyen, Q.~M.~K. Le, D.~K. Nguyen and C.~T. Nguyen are with the Vietnamese-German University,
Vietnam (e-mail of V.~M.~K. Nguyen: 10423063@student.vgu.edu.vn). L.~T.~P. Nguyen is with the University of North Carolina
at Chapel Hill, Chapel Hill, NC, USA. D.~T. Chung is with Florida International University, Miami, FL, USA.
H.~T. Nguyen is with Carnegie Mellon University, Pittsburgh, PA, USA. M.~H.~N. Le is with the Yale School of Medicine,
New Haven, CT, USA. P.~K. Huynh is with North Carolina A\&T State University, Greensboro, NC, USA. T.~H. Nguyen and
U. Bagci are with Northwestern University, Chicago, IL, USA.}}

\maketitle

\begin{abstract}

Alzheimer's disease is diagnosed from clinical, biospecimen and imaging evidence, but few patients undergo every
test. Because these modalities are complementary, a missing one has many plausible values. Most existing methods
leave it out or train on one estimate, so their classifier never sees its other values, and the estimate is never
checked against the diagnosis. We address both gaps with on-policy distillation: the model fills in the missing
modalities itself, and teachers with information available only in training grade these fills. Our method, Graded
Rollout Imputation with Privileged teachers, draws several fills, one modality at a time. An imputation teacher that
knows the diagnosis grades each step, bringing the outcome into the imputation as statistical practice recommends. A
diagnosis teacher that sees all real modalities supplies a soft target. Trained on fills that vary, the model learns a
confidence that reflects what the missing tests leave open, and it diagnoses from a single rollout. On the
Alzheimer's Disease Neuroimaging Initiative cohort, it outperforms state-of-the-art baselines. On two external cohorts, it remains competitive in discrimination and, even after every
method is recalibrated, its calibration error is level with the lowest on both. Added to three of these baselines, the
teachers improve each of them. Our method offers a single, well-calibrated model for whatever combination of tests a
patient has received.

\end{abstract} 


\section{Introduction}
\label{sec:intro}

\begin{figure}[!t]
\centering
\includegraphics[width=0.92\columnwidth]{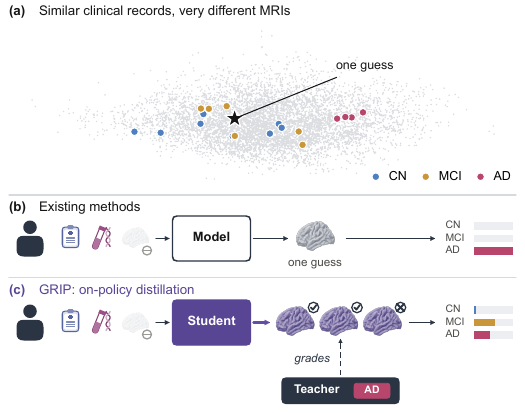}
\caption{A missing test has no single right answer. (a) Regional MRI measurements of ADNI training visits (gray,
first two principal components) and of the 20 visits whose clinical records are most similar to one visit
(colored by diagnosis). The star is a single estimate of that visit's MRI from its clinical record. (b) Existing
methods train on one estimate. (c) GRIP draws several fills, and a teacher that knows the diagnosis grades each
step of the rollout. Rows (b) and (c) are schematic.}
\label{fig:motivation}
\end{figure}

\begingroup\renewcommand{\thefootnote}{}\footnotetext{\sffamily This work has been submitted to the IEEE for possible publication. Copyright may be transferred without notice, after which this version may no longer be accessible.}
\footnotetext{\sffamily This project is partially supported by NIH
R21-HL186024 (\mbox{SPARTACUS}) grant.}\endgroup
\IEEEPARstart{A}{lzheimer's} disease (AD) is a progressive neurodegenerative disease and the most common cause of
dementia, contributing to an estimated 60--70\% of the 57 million people living with dementia worldwide
\cite{jack2024revised,who2025dementia}. Biospecimen markers (B), such as CSF amyloid and tau, show
whether AD pathology is present but not how far the disease has progressed, and structural MRI (I) shows how much the
brain has degenerated but not whether AD caused it \cite{jack2024revised}. Clinical records (C), such as age, medical
history and neurological examination, measure neither, but are routinely collected and add context to both. A reliable diagnosis, whether a
person is cognitively normal (CN), has mild cognitive impairment (MCI) or has AD, therefore combines the three.
Multimodal models that do so diagnose AD more accurately than models built on any single modality
\cite{venugopalan2021multimodal,castellano2024multimodal,duenias2025hyperfusion}.

Yet these models assume that every patient has undergone every test, which rarely holds in clinical practice. MRI is
costly and not available everywhere, a lumbar puncture is invasive, and the tests a patient receives depend on
the clinic's protocol \cite{mattsson2024plasma,yang2026unimix,yang2026prapoe}. Each patient
therefore arrives with a different subset of modalities, a setting known as \emph{arbitrary modality missingness}
\cite{yang2026unimix,yang2026prapoe}. Methods for this setting combine only the modalities that are present
\cite{feng2024anymod,yun2024flexmoe}, first estimate the missing ones
\cite{yoon2018gain,wang2023shaspec,zhang2022m3care,yang2026unimix,yang2026prapoe}, or distill a teacher trained on all
modalities into a student that sees fewer \cite{hu2020kdnet,wei2023mmanet,guan2021mri,kwak2025icmkd,liu2025agmd,wang2026e2ad}.

Because the modalities are complementary, the observed ones cannot pin down a missing one: patients with similar
clinical records can have very different MRIs (Fig.~\ref{fig:motivation}a), as clinical presentation and brain structure
are only loosely coupled \cite{stern2012reserve}. A missing modality has many plausible
values rather than a single right one. Most methods leave it out, replace it with a placeholder, or estimate it once,
and train their classifier on that one value (Fig.~\ref{fig:motivation}b), although a single imputed value ignores how uncertain the missing
value is \cite{sterne2009multiple}. The classifier then never sees the other values the missing modality could
take, and its confidence cannot reflect them. PRA-PoE down-weights uncertain estimates so that they do not make the model overconfident, but it still
synthesizes only one value of each missing modality \cite{yang2026prapoe}.

Nor is any estimate checked for whether it is plausible for the patient's diagnosis, even in methods
that sample several
\cite{ma2021smil}. The missing values depend on the diagnosis: the
missing MRI of a patient with AD should show more atrophy than that of a cognitively normal patient
\cite{jack2024revised}. This is why multiple imputation conditions its imputation model on the outcome
\cite{sterne2009multiple,moons2006using}. A diagnostic model cannot do this directly, because the diagnosis is what it
must predict. Existing methods therefore train their estimates only through the final diagnosis
\cite{wang2023shaspec,zhang2022m3care,yang2026unimix,yang2026prapoe}, to look like real data \cite{yoon2018gain}, or to
match the patient's own real data \cite{gao2022tpagan,pan2022dsnet,hoffman2016hallucination}. Distillation-based methods supervise the student only on the inputs it is given, never on its own
outputs, which is known as \textbf{off-policy distillation} \cite{yin2026hopd}, and in AD they need one student per
pattern of missing modalities \cite{guan2021mri,kwak2025icmkd,liu2025agmd,wang2026e2ad}. None of them asks
whether what the model itself fills in is plausible for the patient's diagnosis, although at test time the
diagnosis rests on it.

\textbf{On-policy distillation}, developed for language models \cite{agarwal2024onpolicy,gu2024minillm} and so
far applied to students that generate text, offers the
missing piece: the student generates its own output, and a teacher that knows more grades it against a whole
distribution rather than a single correct answer. For missing modalities, such a teacher exists during training: for every training patient, the
diagnosis and all real modalities are known. This information, available during training but not at test time, is called \emph{privileged}
\cite{vapnik2009lupi,lopezpaz2016unifying}. A privileged teacher can use the diagnosis to grade the model's own
estimates, much like the outcome-aware imputation model of multiple imputation, while the student, the only model used
at test time, never sees it.

\begin{figure*}[t]
\centering
\includegraphics[width=0.80\textwidth]{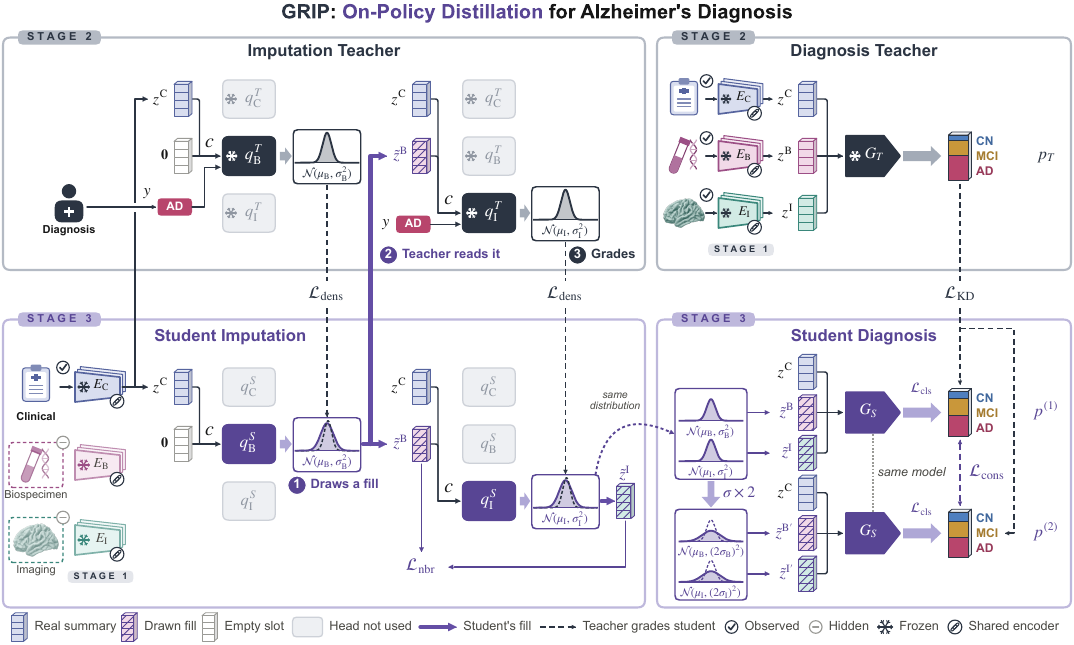}
\caption{Overview of GRIP on one training example: a visit with all three modalities and an AD diagnosis, in a view
that hides B and I. (1) The student draws a fill for B. (2) The frozen imputation teacher, which knows the diagnosis,
reads the student's context for I, which holds this fill. (3) It grades each rollout step ($\mathcal{L}_{\mathrm{dens}}$),
and $\mathcal{L}_{\mathrm{nbr}}$ ties each fill to the patient's own data. The classifier diagnoses two rollouts, the
second wider, learns from the label and the frozen diagnosis teacher ($\mathcal{L}_{\mathrm{cls}}$, with
$\mathcal{L}_{\mathrm{KD}}$) and must agree across them ($\mathcal{L}_{\mathrm{cons}}$). Curves and outputs are schematic.}
\label{fig:overview}
\end{figure*}

We propose Graded Rollout Imputation with Privileged teachers (GRIP, Fig.~\ref{fig:overview}), which turns this
privileged information into two teachers. The student draws several fills of the missing modalities, one modality at
a time, and an \textbf{imputation teacher} that knows the diagnosis grades each step of this rollout (Fig.~\ref{fig:motivation}c). With some
modalities hidden, the hard label still demands a fully confident diagnosis, so a \textbf{diagnosis teacher}, which
sees all of the patient's real modalities, adds a soft diagnosis that says how clear-cut the case is. Trained on fills
that vary, GRIP learns a confidence that reflects what the missing tests leave open, and a single rollout suffices
at test time. This
matters in the clinic, because poorly calibrated predictions can mislead clinical decisions
\cite{vancalster2019calibration}. The teachers are part of how the model is trained, not of the network itself, so
they can be added to other models as well.

We make three contributions:
\begin{itemize}
\item We reframe diagnosis from incomplete modalities as on-policy distillation and propose GRIP, in which an
imputation teacher that knows the diagnosis grades each step of the model's own rollouts and a diagnosis teacher that
sees all real modalities supplies soft targets. To our knowledge, GRIP is the first method to grade its own imputations
by whether they are plausible for the patient's diagnosis, and training on several graded fills, instead of one mean
estimate, improves both accuracy and calibration.
\item We show that the teachers are not tied to GRIP: added to three recent methods, PRA-PoE, UniMIX-AD and
MoE-Retriever, they significantly improve the accuracy and AUC of every one of them.
\item On the Alzheimer's Disease Neuroimaging Initiative (ADNI) cohort \cite{petersen2010adni}, GRIP outperforms
state-of-the-art baselines. On two external cohorts,
AIBL and OASIS-3, it remains competitive in discrimination and, even after every method is recalibrated, its
calibration error is level with the lowest on both, about half or less than that of every deep baseline
on AIBL.
\end{itemize}

\section{Related Work}
\label{sec:related}

\subsection{Multimodal and incomplete multimodal learning for AD}
Multimodal methods for AD fuse MRI with amyloid PET
\cite{castellano2024multimodal}, or with genetic, clinical and CSF data, by concatenation or by conditioning of the
imaging network \cite{venugopalan2021multimodal,duenias2025hyperfusion}. These models, like multimodal fusion networks in other neuroimaging tasks
\cite{chung2026modality}, are built for patients who have a fixed set of modalities.

Generative methods synthesize a missing modality and then diagnose, for example PET from MRI with GANs
\cite{gao2022tpagan}, in one case guided by the features of a diagnosis network \cite{pan2022dsnet}. They need extra
synthesis networks \cite{jia2026mdsfd}, can add noise to the diagnosis \cite{chen2023dftd,kwak2025icmkd}, and mostly
follow a fixed path such as MRI to PET, so covering every subset of modalities would take a combinatorial number of
models \cite{yang2026unimix}. Distillation methods avoid synthesis at test time: a teacher trained on several
modalities guides a student that sees fewer, such as an MRI-only student
\cite{guan2021mri,liu2025agmd,kwak2025icmkd,wang2026e2ad}, and some also estimate features of the missing modality
\cite{chen2023dftd,jia2026mdsfd}. Outside AD, KD-Net and MMANet distill a complete-modality teacher into an
incomplete-modality student \cite{hu2020kdnet,wei2023mmanet}. Each student, however, handles only one pattern of
missing modalities.

Methods for arbitrary modality missingness, where one model must diagnose from any subset of modalities, align
representations across all combinations (AnyMod \cite{feng2024anymod}), combine a missing-modality bank with experts
for each combination (Flex-MoE \cite{yun2024flexmoe}), reconstruct missing representations by cross-attention
(UniMIX-AD \cite{yang2026unimix}) or re-synthesize missing features and down-weight uncertain ones in a product of
experts (PRA-PoE \cite{yang2026prapoe}). They leave a missing modality out or make one estimate of it, and none grades
its estimate for whether it is plausible for the patient's diagnosis.

\subsection{Multiple imputation and on-policy distillation}
Statistics answers the uncertainty of a missing value with multiple imputation, which fills it several times,
analyzes each filled dataset and pools the results \cite{sterne2009multiple}. Its imputation model should include the
outcome \cite{sterne2009multiple,moons2006using}. In prediction, however, the outcome is unknown when a new patient is
diagnosed. Deep generative models also draw several values of what is missing \cite{shi2019mmvae}, but without the
diagnosis. GRIP keeps the several draws and brings in the outcome through a teacher used only in training.

The distillation methods above are off-policy \cite{yin2026hopd}, and their teachers hold privileged information
\cite{vapnik2009lupi,lopezpaz2016unifying}, as in modality hallucination \cite{hoffman2016hallucination}. Beyond
language models, H-OPD combines a vision-language teacher and text-only teachers at each token of the student's output
\cite{yin2026hopd}. In GRIP, the student generates fills of the missing modalities instead of text, and a teacher
privileged by the diagnosis grades each step of its rollouts.

\section{Methodology}
\label{sec:method}

\subsection{Overview}
GRIP is trained in three stages (Fig.~\ref{fig:overview}). Stage~1 trains one encoder per modality, which compresses
it into a short summary. Stage~2 trains the two \emph{teachers}: the \emph{imputation teacher}, which knows the
diagnosis and predicts what a missing summary looks like, and the \emph{diagnosis teacher}, which sees all of a
patient's real modalities and diagnoses the visit. Stage~3 trains the \emph{student}, the only model used at test time,
which fills each missing summary with a random draw and then diagnoses.

A \emph{fill} is a drawn summary of a missing modality. A \emph{view} is a training copy of a visit in which some of
its real modalities are hidden. Fig.~\ref{fig:overview} follows one view through the three stages.

\subsection{Problem definition}
Each visit $i$ has up to $M=3$ modalities $x_i^{m}$, $m\in\{\mathrm{C},\mathrm{B},\mathrm{I}\}$, an availability mask
$r_i\in\{0,1\}^{3}$ and a three-class diagnosis $y_i\in\{\mathrm{CN},\mathrm{MCI},\mathrm{AD}\}$. At test time, the
model receives any non-empty subset $S$ of the available modalities and must predict $y_i$.

\subsection{Stage 1: label-free modality encoders}
\label{sec:enc}
Each modality has its own encoder, $E_{\mathrm{C}}$, $E_{\mathrm{B}}$ or $E_{\mathrm{I}}$, which compresses it into a
short summary vector $z^{m}=E_m(x^{m})$. It is trained to reconstruct the modality from its summary, without the
diagnosis, so the summaries cannot carry memorized training labels into the later stages. The encoders are then
frozen, and everything that follows works on the summaries $z^{\mathrm{C}}$, $z^{\mathrm{B}}$ and $z^{\mathrm{I}}$.

\subsection{Density heads}
\label{sec:density}
GRIP fills a missing modality at the level of its summary $z^{m}$. Since this summary is uncertain, GRIP predicts a distribution of plausible summaries rather than a single vector.

For a patient with only clinical data, the biospecimen summary (B) is filled from $z^{\mathrm{C}}$, and the
imaging summary (I) then from $z^{\mathrm{C}}$ and the fill $\tilde z^{\mathrm{B}}$ (Fig.~\ref{fig:overview}). Each step thus conditions on what is known about
the other two modalities, which we call the \emph{context} $c$ of the missing modality $m$: one slot for each other
modality $m'$,
\begin{equation}
\mathrm{slot}(m')=
\begin{cases}
z^{m'}, & \text{if } m' \text{ is observed},\\
\tilde z^{m'}, & \text{if } m' \text{ is already filled},\\
\mathbf{0}, & \text{otherwise},
\end{cases}
\end{equation}
each with a bit marking whether it is empty. A \emph{density head}, one per modality, maps the context to a Gaussian
distribution over the missing summary:
\begin{equation}
q_m(z^{m}\mid c)=\mathcal{N}\big(\mu(c),\operatorname{diag}\sigma^{2}(c)\big),
\end{equation}
where the mean $\mu(c)$ is the most likely summary and the standard deviation $\sigma(c)$ the imputation uncertainty.
Both the student and the imputation teacher use density heads of this form. Only the teacher's heads also receive the
diagnosis $y$.

\subsection{Stage 2: two privileged teachers}
\label{sec:teacher}
During training, two pieces of information are available that the student never receives: each patient's diagnosis
$y$ and all of the modalities the patient actually has. Stage~2 trains one teacher on each.

\emph{Imputation teacher.} Knowing the diagnosis narrows the plausible values of a missing modality. The imputation teacher is
therefore a density head that receives the diagnosis $y$ in addition to the context $c$:
\begin{equation}
q^{T}_m(z^{m}\mid c,y)=\mathcal{N}\big(\mu_T(c,y),\operatorname{diag}\sigma_T^{2}(c,y)\big).
\end{equation}
It is the outcome-aware imputation model that multiple imputation recommends
\cite{sterne2009multiple,moons2006using}. The teacher is trained only on visits where modality $m$ was actually
acquired. We hide its true summary $z^m$ as the target, and the remaining real modalities, in every non-empty subset,
form the context. The head minimizes the negative log-likelihood of the real summary:
\begin{equation}
\mathcal{L}^{T}_{m}=-\log q^{T}_m(z^{m}\mid c,y).
\end{equation}

\emph{Diagnosis teacher.} Diagnostic labels are discrete, yet AD progresses on a continuum. A one-hot label hides
how borderline a case is, whereas a classifier that observes all of a patient's real modalities can tell clear-cut cases from borderline ones. The diagnosis teacher is such a classifier, $G_T$, trained with cross-entropy across all non-empty subsets of
each patient's available modalities. Its prediction from all observed modalities,
\begin{equation}
p_T=\operatorname{softmax}\big(G_T(z)\big),
\end{equation}
is a soft diagnosis of the visit.

\emph{Out-of-fold teachers.} A teacher applied to patients it was trained on would collapse toward the one-hot label,
and its soft target would add nothing. Both teachers are therefore trained out-of-fold: the training participants are
partitioned into five person-disjoint parts, and for each part both teachers are trained on the other four. The
held-out part is used only to decide when to stop training. The teachers are frozen after Stage~2.

\subsection{Stage 3.1: the student}
\label{sec:student}
The student, the deployed model, consists of unprivileged density heads $q^{S}_m(z^{m}\mid c)=\mathcal{N}\big(\mu_S(c),\operatorname{diag}\sigma_S^{2}(c)\big)$,
which fill missing modalities from the available context, and a classifier $G_S$, which predicts the diagnosis. The
classifier receives the three summaries (real or filled), a bit per slot marking whether it was filled, and the mean
standard deviation of each fill, so it knows which inputs were drawn and how uncertain they are.

\emph{Training views.} To cover every missingness pattern, the student is trained on every non-empty subset of a
visit's real modalities, which gives up to seven views for a complete visit. Trained with the label alone, these
views amount to standard modality dropout, our baseline (Section~\ref{sec:ablation}). In GRIP, the two teachers
supervise every view.

\emph{Rollouts.} For a view with missing modalities, the student fills every slot the view hides, including
modalities the visit never had, one modality at a time, each fill drawn from the student's density head for that modality:
\begin{equation}
\tilde z^{m}=\mu_S(c)+\gamma\,\sigma_S(c)\odot\epsilon,\qquad \epsilon\sim\mathcal{N}(0,I),
\label{eq:rollout}
\end{equation}
where the tilde marks a fill and $\gamma$ scales the spread of the draw. Each fill joins the context of the next step,
so a later fill depends on the earlier ones, and we call this sequence of draws a \emph{rollout}. By default $\gamma=1$,
and a second, wider rollout is introduced in Section~\ref{sec:gri}. When two modalities are missing, the rollout is
drawn in both orders, each with weight one half. Because rollouts are drawn the same way during training and
inference, the classifier is trained on the kind of input it receives at test time.

\subsection{Stage 3.2: graded rollouts}
\label{sec:gri}
A random draw helps only if it is plausible and fits the patient. Three losses ensure this: the imputation teacher
grades each step of the rollout, each fill is aligned with the patient's real data, and the classifier learns not to
depend on any single draw.

\emph{Grading by the imputation teacher.} At each rollout step, the imputation teacher reads the student's context and
pulls the student's density toward its own through the reverse Kullback--Leibler (KL) divergence
\cite{gu2024minillm}:
\begin{equation}
\mathcal{L}_{\mathrm{dens}}=\sum_{m\ \mathrm{filled}}\mathrm{KL}\!\left(q^{S}_m(\cdot\mid c)\,\middle\|\,
q^{T}_m(\cdot\mid c,y)\right).
\end{equation}
Both densities are Gaussian, so the divergence has a closed form. The grading is on-policy in the sense of language
models \cite{agarwal2024onpolicy,gu2024minillm}: the teacher judges the student where the student's own rollout takes
it. When a view hides two modalities, the second fill is drawn from a context that holds the student's first fill,
and the teacher grades the student at this context rather than at one built from real data. When only one modality
is missing, the context holds real data only, and the two ways of grading are the same.

\emph{What the student learns from the teacher.} The student does not see $y$, so it is trained toward the teacher
averaged over the diagnoses that occur with each context. The fills are thus plausible across the diagnoses the context
allows, and the classifier is left to tell which diagnosis it is.

\emph{Neighborhood alignment.} The imputation teacher makes a fill plausible for the disease stage, but not
necessarily for the individual. When a view hides an observed modality, its true summary $z^m$ is available. Forcing
the fill $\tilde z^m$ to match $z^m$ would remove the diversity of the fills, so we ask instead that $\tilde z^m$ have
the same neighbors as $z^m$. Let $\pi$ and $\tilde\pi$ be the softmax distributions over the cosine similarities of
$z^m$ and $\tilde z^m$ to the real summaries of the other patients in the mini-batch, at temperature $\tau_n$. We align
the two neighborhoods, following stochastic neighbor embedding \cite{hinton2002sne}:
\begin{equation}
\mathcal{L}_{\mathrm{nbr}}=\mathrm{KL}(\tilde\pi\,\|\,\pi).
\end{equation}

\emph{Two-rollout consistency.} Different plausible draws should not change the diagnosis. For every view, a second
rollout is drawn with a wider spread ($\gamma>1$ in Eq.~\eqref{eq:rollout}), and the two predictions $p^{(1)}$ and
$p^{(2)}$ are asked to agree,
\begin{equation}
\mathcal{L}_{\mathrm{cons}}=\tfrac12\big[\mathrm{KL}(p^{(1)}\|p^{(2)})+\mathrm{KL}(p^{(2)}\|p^{(1)})\big],
\end{equation}
following R-Drop \cite{liang2021rdrop} with drawn fills in place of dropout masks. The classifier thus relies on what
the plausible fills share, and the wider second rollout also shows it less likely fills.

\subsection{Stage 3.3: soft target from the diagnosis teacher}
\label{sec:psd}
On a weak view, such as B only, the label demands a confident diagnosis that the data may not support. Whenever a
training view hides modalities that the patient actually had, the diagnosis teacher, which read all of them, provides
a soft target $p_T$ that says how borderline the case is.

With student predictions $p_S=\operatorname{softmax}\big(G_S(\tilde z)\big)$, the classification objective combines
the label with distillation from the diagnosis teacher:
\begin{equation}
\mathcal{L}_{\mathrm{cls}}=\mathrm{CE}\big(G_S(\tilde z),y\big) + \underbrace{\beta\,\mathrm{KL}\big(p_T\,\|\,p_S\big)}_{\mathcal{L}_{\mathrm{KD}}},
\end{equation}
where $\tilde z$ denotes the view's summaries after the rollout and $\beta$ is a balancing weight. With two rollouts,
$\mathcal{L}_{\mathrm{cls}}$ is averaged over both. This adapts complete-to-incomplete distillation
\cite{hu2020kdnet,wei2023mmanet,lopezpaz2016unifying}. Because the hidden modalities are filled by rollouts, the
classifier receives the teacher's target on the same stochastic inputs it sees at test time.

\subsection{Training objective and inference}
\label{sec:objective}
\emph{Objective.} The overall objective, averaged over the views of a mini-batch, is
\begin{equation}
\mathcal{L}=\mathcal{L}_{\mathrm{cls}}+\mathcal{L}_{\mathrm{cons}}+\mathcal{L}_{\mathrm{dens}}
+\lambda_n\mathcal{L}_{\mathrm{nbr}},\qquad \lambda_n=1.
\label{eq:obj}
\end{equation}
Each loss trains one part of the student: $\mathcal{L}_{\mathrm{dens}}$ and $\mathcal{L}_{\mathrm{nbr}}$ its density
heads, and $\mathcal{L}_{\mathrm{cls}}$ and $\mathcal{L}_{\mathrm{cons}}$ its classifier (the fills enter the
classifier without gradient). After the classifier's best epoch, it is frozen and the density heads are refined for a
few more epochs with more weight on $\mathcal{L}_{\mathrm{nbr}}$, which ties the fills more closely to the individual
patient.

\emph{Inference.} Given any subset of modalities, the encoders summarize the given modalities, the student's density
heads fill the missing ones by a single rollout (in a random order when two are missing), and the classifier diagnoses
from it in one pass. No teacher, label or hidden modality is used.

\section{Dataset and Experimental Setup}
\label{sec:setup}
\subsection{Dataset and features}
\label{sec:data}
We conduct experiments on the Alzheimer's Disease Neuroimaging Initiative (ADNI) \cite{petersen2010adni}, a
multicenter cohort that collects clinical, biospecimen and imaging data at baseline and follow-up visits. Data used in
the preparation of this article were obtained from the Alzheimer's Disease Neuroimaging Initiative (ADNI) database
(\url{adni.loni.usc.edu}). The ADNI was launched in 2003 as a public-private partnership, led by Principal
Investigator Michael W. Weiner, MD. The primary goal of ADNI has been to test whether serial magnetic resonance
imaging (MRI), positron emission tomography (PET), other biological markers, and clinical and neuropsychological
assessment can be combined to measure the progression of mild cognitive impairment (MCI) and early Alzheimer's
disease (AD). The data were downloaded on 12 September 2026 (MUSE regional volumes: 18 October 2026). To assess generalization to independent cohorts, we additionally use the Australian Imaging, Biomarkers and
Lifestyle study (AIBL) \cite{ellis2009aibl}\footnote{Data was collected by the AIBL study group. AIBL study methodology
has been reported previously~\cite{ellis2009aibl}. We use AIBL Data extract 3.3.0 (tables dated 1 June 2018).} and the
Open Access Series of Imaging Studies (OASIS-3) \cite{lamontagne2019oasis3} as external test sets, which are not used
for training or model selection. The task is three-class classification of cognitively normal (CN), MCI and AD, using the clinical
diagnosis at each visit. ADNI contains 16{,}247 visits of 3{,}685 participants (6{,}603/6{,}616/3{,}028 CN/MCI/AD visits,
mean age 74.8 years), AIBL 1{,}674 visits of 856 participants (1{,}226/246/202) and OASIS-3 8{,}407 visits of
1{,}362 participants (6{,}438/601/1{,}368). Clinical data exist at every ADNI visit, 70.1\% of the visits have all
three modalities, and C+I is rare (2.2\%).

From ADNI, we use three modalities. The clinical modality (C) holds demographics, medical history, vital signs,
neurological examination and medication classes. Medication names were assigned to 19 drug classes by a language
model that followed a written labeling guide, with manual corrections, and anti-dementia and trial drugs were
removed. The biospecimen modality (B) holds apolipoprotein E (APOE)
genotype, measured once per participant and carried to all of their visits, and cerebrospinal fluid (CSF) A$\beta_{42}$, tau and p-tau,
available at 24.3\% of the visits. The imaging modality (I) holds FreeSurfer
\cite{fischl2012freesurfer} regional measurements of T1-weighted MRI (volumes, surface areas and cortical thickness).
All methods work on these measurements rather than on the images. Cognitive and functional test scores are excluded. The clinical diagnosis in ADNI is made from them, so they
would leak the label \cite{petersen2010adni,yang2026unimix}.  AIBL shares with ADNI age, sex, medical history,
APOE genotype and T1-weighted MRI, for which MUSE regional volumes \cite{doshi2016muse} are used. OASIS-3 shares
demographics, medical history, vital signs, medication classes, APOE genotype and FreeSurfer regional measurements of
T1-weighted MRI. OASIS-3 records the clinician's diagnosis as text, and uncertain dementia, questionable
impairment, impairment in memory only or incipient dementia, with no other cause named, are mapped to MCI. This MCI
class, impaired but not demented, is broader than a formal MCI diagnosis.

The ADNI participants are split into a fixed held-out test set (20\%, 737 participants) and five cross-validation
folds of the remaining 2{,}948, stratified by ADNI entry phase and last diagnosis, so that no participant appears in
more than one set. Each fold serves once as the validation set, and the other four form the training set. All three
cohorts obtained ethics approval and informed consent at their sites, and we use their de-identified data under the
cohorts' data use agreements.

\subsection{Data pre-processing}
\label{sec:prep}
CSF samples and MRI scans are assigned to the closest diagnostic visit within 90 days, and a modality with no record
in that window is treated as missing. Clinical forms are taken from the most recent record before the visit, or from
one up to 90 days after it, and age is computed at the visit, so every visit has clinical data. Continuous inputs are
z-score normalized and categorical inputs are one-hot encoded. A missing value within an available modality, such as
CSF when only APOE is available, is replaced by the training mean and flagged by a missingness indicator.
Normalization statistics are computed on the training folds only and applied unchanged to the test set and the
external cohorts.

\begin{table*}[t]
\centering
\caption{Performance comparison on the ADNI test set. Values are mean (std) over 15 runs (\%). Simulated: each of
the seven combinations of clinical (C), biospecimen (B) and imaging (I) data, with the other modalities hidden. Real:
each test visit with the modalities actually acquired. ECE: expected calibration error as trained, before any post-hoc calibration. Best result in \textbf{bold}, second
best \underline{underlined}. Superscripts on Mean and Real: the difference from GRIP is
significant in both tests of Section~\ref{sec:protocol}, $^{*}p<0.05$, $^{**}p<0.01$.}
\label{tab:main}
\footnotesize
\setlength{\tabcolsep}{3.1pt}
\begin{tabular}{@{}llcccccccc>{\hspace{8.8pt}}c<{\hspace{8.9pt}}@{}}
\toprule
 & & \multicolumn{8}{c}{Simulated missingness} & Real \\
\cmidrule(lr){3-10}\cmidrule(lr){11-11}
 & & \multicolumn{3}{c}{Single} & \multicolumn{3}{c}{Double} & Complete & & missingness \\
\cmidrule(lr){3-5}\cmidrule(lr){6-8}\cmidrule(lr){9-9}
Metric & Method & C & B & I & C+B & C+I & B+I & C+B+I & Mean & \\
\midrule
\multirow{10}{*}{ACC} & XGBoost \cite{chen2016xgboost} & \textbf{51.5(1.1)} & \underline{45.6(0.4)} & 55.6(0.9) & \underline{53.6(1.0)} & 59.4(1.2) & 56.4(0.8) & 59.7(0.9) & 54.5(0.6) & \underline{58.7(0.8)} \\
 & Flex-MoE \cite{yun2024flexmoe} & 49.8(1.0) & 44.4(0.9) & 53.6(1.6) & 51.3(1.5) & 55.8(1.5) & 53.7(1.3) & 56.1(1.0) & 52.1(0.8)\rlap{$^{**}$} & 55.9(1.0)\rlap{$^{**}$} \\
 & UniMIX-AD \cite{yang2026unimix} & 50.5(1.0) & 44.4(1.1) & 53.4(1.4) & 51.7(1.1) & 56.0(1.1) & 54.5(1.6) & 56.9(1.4) & 52.5(0.7)\rlap{$^{**}$} & 56.0(0.9)\rlap{$^{**}$} \\
 & PRA-PoE \cite{yang2026prapoe} & 49.8(1.3) & \textbf{45.9(0.6)} & 56.8(1.7) & 51.0(1.0) & 59.2(1.8) & \underline{57.9(1.7)} & 59.6(1.7) & 54.3(1.2)\rlap{$^{*}$} & 58.0(1.3)\rlap{$^{*}$} \\
 & ShaSpec \cite{wang2023shaspec} & 49.2(1.9) & 44.2(1.3) & 54.0(1.8) & 50.2(1.6) & 57.8(1.7) & 55.5(1.7) & 58.5(1.8) & 52.8(1.2)\rlap{$^{**}$} & 56.9(1.6)\rlap{$^{**}$} \\
 & M3Care \cite{zhang2022m3care} & 50.7(0.8) & 44.7(1.1) & 56.8(1.2) & 51.5(0.5) & 58.6(1.1) & 55.8(1.2) & 58.3(1.4) & 53.8(0.7)\rlap{$^{**}$} & 57.0(1.0)\rlap{$^{**}$} \\
 & AnyMod \cite{feng2024anymod} & \underline{51.0(1.7)} & 45.3(1.0) & 56.8(1.0) & 52.3(1.9) & \underline{60.0(1.4)} & 57.7(1.2) & \underline{60.3(1.3)} & \underline{54.8(1.0)} & 58.6(1.0) \\
 & MoE-Retriever \cite{yun2025moeretriever} & 48.0(1.0) & 42.9(0.9) & 53.1(0.8) & 49.7(1.3) & 55.1(1.2) & 53.4(1.0) & 55.2(1.3) & 51.0(0.7)\rlap{$^{**}$} & 54.5(1.0)\rlap{$^{**}$} \\
 & MMIN \cite{zhao2021mmin} & \underline{51.0(0.8)} & 45.3(0.3) & \textbf{57.6(1.2)} & 52.0(0.9) & 59.7(1.1) & \underline{57.9(1.5)} & 60.0(1.5) & \underline{54.8(0.8)} & 58.3(1.1) \\
\rowcolor{gray!15}  & GRIP (ours) & \textbf{51.5(0.8)} & 44.4(0.9) & \underline{57.5(0.9)} & \textbf{53.7(0.7)} & \textbf{61.0(1.0)} & \textbf{58.4(0.9)} & \textbf{61.3(1.1)} & \textbf{55.4(0.5)} & \textbf{59.6(0.7)} \\
\midrule
\multirow{10}{*}{AUC} & XGBoost \cite{chen2016xgboost} & \textbf{68.3(0.7)} & \underline{61.7(0.4)} & 74.4(0.7) & \underline{71.3(0.7)} & 77.6(0.6) & 75.2(0.6) & 78.0(0.6) & 72.4(0.4)\rlap{$^{*}$} & 76.8(0.5)\rlap{$^{*}$} \\
 & Flex-MoE \cite{yun2024flexmoe} & 65.9(0.7) & 60.2(0.9) & 72.3(1.0) & 68.9(1.0) & 75.1(1.2) & 72.8(1.4) & 75.0(1.0) & 70.0(0.7)\rlap{$^{**}$} & 73.7(0.9)\rlap{$^{**}$} \\
 & UniMIX-AD \cite{yang2026unimix} & 66.0(0.8) & 61.2(0.3) & 71.9(1.2) & 69.2(0.8) & 74.4(1.0) & 73.0(1.3) & 75.2(1.3) & 70.1(0.7)\rlap{$^{**}$} & 73.8(1.2)\rlap{$^{**}$} \\
 & PRA-PoE \cite{yang2026prapoe} & 65.4(0.9) & \textbf{62.0(0.5)} & 75.2(1.4) & 68.6(1.0) & 77.4(1.3) & 75.9(1.4) & 78.1(1.3) & 71.8(1.0)\rlap{$^{**}$} & 76.0(1.1)\rlap{$^{**}$} \\
 & ShaSpec \cite{wang2023shaspec} & 64.3(1.1) & 61.0(0.7) & 72.6(1.3) & 67.3(1.6) & 76.0(1.4) & 74.0(1.4) & 76.8(1.5) & 70.3(1.0)\rlap{$^{**}$} & 74.6(1.4)\rlap{$^{**}$} \\
 & M3Care \cite{zhang2022m3care} & 65.9(0.6) & 60.3(1.1) & 74.7(0.8) & 67.7(0.6) & 77.2(0.9) & 74.6(0.9) & 77.4(1.0) & 71.1(0.5)\rlap{$^{**}$} & 75.0(0.7)\rlap{$^{**}$} \\
 & AnyMod \cite{feng2024anymod} & 67.2(1.2) & 61.6(0.3) & \underline{75.5(0.8)} & 70.1(1.3) & \underline{78.2(1.1)} & \underline{76.3(0.9)} & \underline{78.7(1.2)} & \underline{72.5(0.9)}\rlap{$^{*}$} & \underline{76.9(1.1)}\rlap{$^{*}$} \\
 & MoE-Retriever \cite{yun2025moeretriever} & 63.4(0.6) & 57.5(1.2) & 71.9(0.9) & 66.5(0.9) & 73.8(1.2) & 72.4(1.0) & 74.1(1.1) & 68.5(0.7)\rlap{$^{**}$} & 71.9(0.9)\rlap{$^{**}$} \\
 & MMIN \cite{zhao2021mmin} & 65.4(1.0) & 61.4(0.4) & 75.2(0.9) & 68.9(0.7) & 77.6(0.9) & 76.0(0.9) & 78.3(0.8) & 71.8(0.6)\rlap{$^{**}$} & 76.2(0.8)\rlap{$^{**}$} \\
\rowcolor{gray!15}  & GRIP (ours) & \underline{67.5(0.5)} & \underline{61.7(0.4)} & \textbf{76.2(0.6)} & \textbf{71.6(0.6)} & \textbf{78.9(0.4)} & \textbf{77.3(0.6)} & \textbf{79.8(0.5)} & \textbf{73.3(0.3)} & \textbf{78.0(0.4)} \\
\midrule
\multirow{10}{*}{ECE} & XGBoost \cite{chen2016xgboost} & 5.6(4.3) & \underline{2.7(1.1)} & 9.2(4.1) & 5.0(4.2) & 8.8(5.0) & 9.5(4.0) & 9.1(4.8) & 7.1(3.8)\rlap{$^{**}$} & 7.3(4.8)\rlap{$^{**}$} \\
 & Flex-MoE \cite{yun2024flexmoe} & 7.3(2.7) & 5.4(3.2) & 16.3(4.2) & 8.2(3.8) & 18.3(4.1) & 18.8(4.3) & 17.6(5.6) & 13.1(3.2)\rlap{$^{**}$} & 14.4(4.9)\rlap{$^{**}$} \\
 & UniMIX-AD \cite{yang2026unimix} & 5.8(2.6) & 4.0(2.2) & 12.1(5.0) & 6.3(2.7) & 11.9(5.7) & 12.7(5.7) & 16.4(6.4) & 9.9(3.6)\rlap{$^{**}$} & 13.0(4.8)\rlap{$^{**}$} \\
 & PRA-PoE \cite{yang2026prapoe} & 10.1(4.5) & 3.6(0.9) & 11.1(6.7) & 9.9(4.6) & 10.7(6.5) & 11.0(6.8) & 9.9(6.7) & 9.5(5.1)\rlap{$^{**}$} & 9.3(6.2)\rlap{$^{**}$} \\
 & ShaSpec \cite{wang2023shaspec} & 15.1(8.6) & 12.4(7.3) & 18.2(7.6) & 16.6(7.9) & 17.5(6.8) & 18.3(7.0) & 18.3(6.6) & 16.6(6.2)\rlap{$^{**}$} & 17.2(7.1)\rlap{$^{**}$} \\
 & M3Care \cite{zhang2022m3care} & 8.2(1.8) & 7.1(1.9) & 10.5(4.9) & 7.2(1.8) & 10.9(4.3) & 9.6(4.3) & 11.4(4.4) & 9.3(2.9)\rlap{$^{**}$} & 9.7(3.5)\rlap{$^{**}$} \\
 & AnyMod \cite{feng2024anymod} & 7.7(6.3) & 4.7(2.7) & 7.9(4.7) & 6.3(6.8) & \underline{6.6(5.8)} & \underline{6.3(5.3)} & \underline{6.1(5.2)} & 6.5(5.0)\rlap{$^{**}$} & 5.9(5.2)\rlap{$^{**}$} \\
 & MoE-Retriever \cite{yun2025moeretriever} & 6.9(2.4) & 4.4(2.7) & 25.3(2.5) & 5.7(2.4) & 21.8(2.1) & 24.2(2.5) & 21.9(2.3) & 15.7(1.7)\rlap{$^{**}$} & 17.0(1.8)\rlap{$^{**}$} \\
 & MMIN \cite{zhao2021mmin} & \underline{4.0(2.0)} & \textbf{2.2(0.7)} & \underline{7.6(5.2)} & \underline{4.0(2.0)} & 7.1(5.3) & 7.3(5.1) & 7.4(5.2) & \underline{5.7(3.3)}\rlap{$^{**}$} & \underline{5.6(4.3)}\rlap{$^{*}$} \\
\rowcolor{gray!15}  & GRIP (ours) & \textbf{3.9(1.0)} & \underline{2.7(1.5)} & \textbf{3.9(1.0)} & \textbf{2.7(0.8)} & \textbf{3.8(0.8)} & \textbf{3.4(0.7)} & \textbf{3.2(1.1)} & \textbf{3.4(0.5)} & \textbf{2.8(1.0)} \\
\bottomrule
\end{tabular}
\end{table*}

\begin{table*}[t]
\centering
\caption{External validation on AIBL and OASIS-3. Values are mean (std) over 15 runs (\%). Both cohorts share APOE
genotype with ADNI but not CSF, so their biospecimen modality is APOE alone. Each method is trained on ADNI with the
inputs the cohort shares with ADNI and tested on all its labeled visits with their acquired modalities. Every method is recalibrated as in Section~\ref{sec:protocol}, which leaves
BACC and CN/AD unchanged. CN/AD: balanced accuracy between CN and AD visits. Best result per column in \textbf{bold}, second best \underline{underlined}.}
\label{tab:aibl}
\footnotesize
\setlength{\tabcolsep}{4pt}
\begin{tabular}{lcccccccc}
\toprule
& \multicolumn{4}{c}{AIBL} & \multicolumn{4}{c}{OASIS-3} \\
\cmidrule(lr){2-5} \cmidrule(lr){6-9}
Method & BACC & AUC & ECE & CN/AD & BACC & AUC & ECE & CN/AD \\
\midrule
XGBoost \cite{chen2016xgboost} & \textbf{50.5(1.4)} & 71.7(1.2) & \textbf{4.2(1.8)} & \underline{79.1(1.2)} & 41.1(1.1) & 65.4(0.6) & \textbf{7.6(1.1)} & 65.8(1.0) \\
Flex-MoE \cite{yun2024flexmoe} & 49.2(2.1) & 72.3(1.1) & 10.4(3.0) & 77.0(2.6) & 40.9(1.1) & 65.7(1.2) & 13.3(3.5) & 64.4(2.0) \\
UniMIX-AD \cite{yang2026unimix} & 50.0(1.8) & 71.9(0.9) & 8.4(2.0) & 78.3(1.7) & 41.8(1.6) & 66.4(0.8) & 14.4(3.8) & 64.5(2.1) \\
PRA-PoE \cite{yang2026prapoe} & 50.1(1.7) & \underline{73.0(1.4)} & 9.4(3.6) & 79.0(1.4) & \textbf{43.5(1.7)} & \underline{68.1(2.9)} & 14.4(4.0) & 66.2(1.5) \\
ShaSpec \cite{wang2023shaspec} & 47.7(2.3) & 70.7(1.8) & 8.6(3.1) & 74.9(2.6) & 40.0(2.3) & 64.4(3.0) & \underline{8.6(4.4)} & 64.4(3.2) \\
M3Care \cite{zhang2022m3care} & \underline{50.4(1.1)} & \textbf{73.6(1.1)} & 14.1(3.0) & 78.9(1.6) & \underline{42.7(1.5)} & \textbf{68.9(1.2)} & 19.8(1.8) & \textbf{66.9(1.0)} \\
AnyMod \cite{feng2024anymod} & 49.8(2.4) & 72.0(2.1) & 9.5(3.0) & 76.5(2.5) & 41.5(2.1) & 65.7(2.3) & 10.8(6.1) & 64.2(2.5) \\
MoE-Retriever \cite{yun2025moeretriever} & 49.2(1.7) & 71.3(1.1) & 9.6(3.1) & 75.8(2.8) & 40.9(1.6) & 64.3(1.1) & 12.0(2.1) & 63.9(1.7) \\
MMIN \cite{zhao2021mmin} & 49.6(1.5) & 72.2(0.8) & \underline{8.2(2.4)} & 77.0(1.3) & 41.5(1.4) & 67.1(1.7) & 10.5(4.3) & 64.9(1.1) \\
\rowcolor{gray!15} GRIP (ours) & 50.2(1.7) & 72.5(0.9) & \textbf{4.2(1.4)} & \textbf{79.6(0.8)} & 42.0(1.0) & 67.2(1.0) & 8.7(2.1) & \underline{66.4(1.1)} \\
\bottomrule
\end{tabular}
\end{table*}

\subsection{Baselines}
\label{sec:baselines}
We compare GRIP with eight recent methods for arbitrary missing modalities, in the order of Table~\ref{tab:main}:
the sparse mixtures of experts Flex-MoE \cite{yun2024flexmoe} and UniMIX-AD \cite{yang2026unimix}, the product of
experts PRA-PoE \cite{yang2026prapoe}, ShaSpec \cite{wang2023shaspec} and M3Care \cite{zhang2022m3care}, which complete
a missing modality from the observed ones or from similar patients, the modality-agnostic Transformer AnyMod
\cite{feng2024anymod}, MoE-Retriever \cite{yun2025moeretriever}, which retrieves a missing embedding from other
patients, and MMIN \cite{zhao2021mmin}, which generates it with cascaded residual autoencoders. We also compare with
XGBoost \cite{chen2016xgboost}, gradient-boosted trees, the standard model for incomplete tabular data, which routes a
hidden modality's missing values by directions learned in training. Distillation methods for AD
\cite{guan2021mri,kwak2025icmkd,liu2025agmd,wang2026e2ad} train one student per fixed set of inputs, so they cannot
diagnose every combination and are not compared directly. Their idea is tested inside GRIP: without the graded
rollouts, GRIP reduces to a teacher that reads all of a visit's modalities supervising a student that sees only some
of them, trained over every combination (``w/o graded rollouts'' in Section~\ref{sec:ablation}).

The baselines run with or are rebuilt from the authors' code, except AnyMod, which has no public code and is
implemented from its paper. Every method keeps its own architecture, losses, optimizer, schedule and the authors'
recommended hyperparameters, and all are trained with GRIP's training views (Section~\ref{sec:student}) and model selection
(Section~\ref{sec:impl}), so that they differ from GRIP only in the method itself.

\subsection{Evaluation protocol and metrics}
\label{sec:protocol}
Each method is trained with three seeds on each of the five
folds, giving fifteen runs, and every run is tested once on the held-out set. To test every pattern of missingness, each
of the seven modality combinations is simulated on the test visits that have all of its modalities, by hiding the
others. The test visits are also evaluated with the modalities actually acquired at each (real missingness). Three of the
seven, B, I and B+I, never occur in ADNI and test robustness to any pattern of missingness, whereas real missingness
covers the combinations that do occur, at their actual frequencies. We report
accuracy (ACC), macro one-vs-rest area under the receiver operating characteristic curve (AUC) and the expected calibration error (ECE) as mean (std) over the fifteen
runs. The ECE measures how well a model's confidence matches its accuracy \cite{guo2017calibration}: the gap between
confidence and accuracy is averaged over fifteen equal-width bins of the top-class probability, weighted by their size,
so lower is better. For the seven combinations, the ECE is averaged over the combinations. To compare calibration on
equal terms, we also recalibrate every method, GRIP included, by temperature scaling \cite{guo2017calibration}
with one temperature per modality combination, which is known at test time, fitted by negative log-likelihood on the
method's own ADNI validation predictions of the same run. Table~\ref{tab:aibl} and the right panels of
Fig.~\ref{fig:reliability} use these recalibrated predictions, whereas Tables~\ref{tab:main},~\ref{tab:plugin} and~\ref{tab:ablation}
report the calibration each model learns in training. The external cohorts keep the temperatures fitted on ADNI,
which tests calibration under the dataset shift they introduce, where post-hoc calibration is known to hold poorly
\cite{ovadia2019trust}. A difference is called
significant only when two tests agree: a paired Wilcoxon signed-rank test over the 15 runs, which reflects training
variability, and a paired bootstrap that resamples the test participants 2{,}000 times with all their visits,
which reflects the choice of test participants.

For each external cohort, every method is retrained on ADNI with only the inputs that cohort shares, and nothing is
tuned on the external cohorts. Both are dominated by CN visits, so we report balanced accuracy (BACC) in place of
accuracy, together with AUC, ECE, and the balanced accuracy between CN and AD visits, where each visit is assigned the
more probable of the two classes.

\subsection{Implementation details}
\label{sec:impl}
GRIP is implemented in PyTorch and trained on four-core CPUs, the deep baselines on NVIDIA T4 GPUs and XGBoost on
an eight-core CPU. Each
modality is encoded into a 128-dimensional summary. The encoders, the density heads and the classifier are small
multilayer perceptrons (MLPs) with one or two hidden layers of 256 units. All networks are trained with Adam \cite{kingma2015adam}, a
learning rate of $10^{-3}$ and a weight decay of $10^{-4}$ for at most 100 epochs, with early stopping. During the
student's first five epochs, later rollout steps read the real summary where it exists. The distillation weight is
$\beta=\ln 3/\mathrm{KL}_0$, where $\mathrm{KL}_0$ is the mean divergence of $p_T$ from the uniform distribution, so
that the distillation term starts at the cross-entropy of a uniform guess. The second rollout uses $\gamma=2$ and the
neighborhood alignment $\tau_n=0.3$. 
The only settings chosen on the validation sets are GRIP's own, such as its dropout rate, $\sigma_{\min}$,
$\gamma$, $\tau_n$ and the refinement weights, chosen one at a time. The model
is selected on the validation score $\tfrac12(\mathrm{acc}_{\mathrm{val}}+\overline{\mathrm{acc}}_{7})$, where
$\mathrm{acc}_{\mathrm{val}}$ is the accuracy on the validation visits with their acquired modalities and
$\overline{\mathrm{acc}}_{7}$ the mean accuracy over the seven combinations. The code will be made publicly available
upon publication.

\section{Results}
\label{sec:results}
\subsection{GRIP outperforms recent methods on ADNI}
\begin{figure}[t]
\centering
\includegraphics[width=0.88\columnwidth]{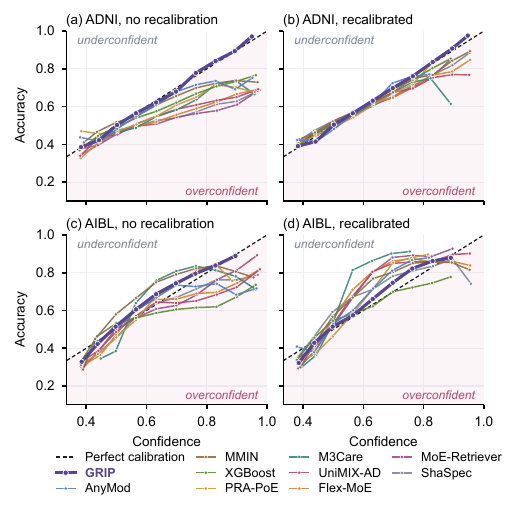}
\caption{Reliability diagrams under real missingness, with the predictions of the 15 runs grouped into 15 confidence
bins. Top: ADNI test set. Bottom: AIBL. Left: no recalibration. Right: every method recalibrated with one
temperature per modality combination, fitted on its own ADNI validation predictions. The baselines are the
reruns that saved validation predictions. Points below the diagonal are overconfident.}
\label{fig:reliability}
\end{figure}

\emph{Overall performance.} GRIP leads every summary column of Table~\ref{tab:main}: the average accuracy, AUC and
ECE over the seven combinations, and all three under real missingness. Its AUC is significantly
higher than that of every baseline and its accuracy than that of six of them, and both vary least across
runs.

\emph{A strong reference.} Our modality-dropout model, the same
encoders and classifier trained on every view with the label alone (``w/o both teachers'' in
Table~\ref{tab:ablation}), is already on par with the best published methods. The teachers significantly improve its
AUC and calibration error and also raise its accuracy, so GRIP's gains add to a strong reference.

\emph{Calibration.} As trained, GRIP's ECE is
significantly lower than that of every baseline, over the seven combinations and under real missingness
(Table~\ref{tab:main}). Under real missingness, its accuracy follows its confidence over the whole range,
whereas every baseline becomes overconfident as its confidence rises (Fig.~\ref{fig:reliability}a).
Recalibration with per-combination temperatures brings every method close to the diagonal on ADNI, and GRIP as trained already matches every recalibrated deep baseline (Fig.~\ref{fig:reliability}b). 

\emph{Across combinations.} GRIP has the highest AUC on five of the seven combinations and is the best method in
all three metrics on the complete combination, the rare C+I and the never-acquired B+I. Against the
strongest baseline, AnyMod, it is ahead on at least six of the seven combinations in every metric (Table~\ref{tab:main}).  

\emph{Efficiency.} The teachers are used only during training, so GRIP is small at inference (0.91\,M parameters)
and needs the fewest floating-point operations of all timed methods, over 80 times fewer than AnyMod (1.13\,M against
94.3\,M per patient).

\begin{table}[t]
\centering
\caption{The two teachers added to three recent methods on the ADNI test set (mean (std) over 15 runs, \%). Each method
runs with its own recipe on the raw inputs, retrained in this experiment, so its values differ slightly from
Table~\ref{tab:main}. ECE of the trained models, before any post-hoc calibration. Better result per method in
\textbf{bold}. $^{*}p<0.05$, $^{**}p<0.01$ versus the method alone in both tests of Section~\ref{sec:protocol}.}
\label{tab:plugin}
\footnotesize
\setlength{\tabcolsep}{2pt}
\begin{tabularx}{\columnwidth}{@{}l*{4}{>{\centering\arraybackslash}X}@{}}
\toprule
 & \multicolumn{3}{c}{Seven combinations} & Real \\
\cmidrule(lr){2-4}\cmidrule(lr){5-5}
Model & ACC & AUC & ECE & ACC \\
\midrule
PRA-PoE & 54.3(0.9) & 72.0(0.5) & 5.1(2.1) & 57.8(1.1) \\
\rowcolor{gray!15} \quad + both teachers & \textbf{55.1(0.7)}$^{*}$ & \textbf{72.6(0.5)}$^{*}$ & \textbf{4.5(1.5)} & \textbf{58.4(0.8)} \\
\addlinespace[2pt]
UniMIX-AD & 52.4(0.6) & 70.2(0.5) & 7.9(2.3) & 56.0(0.9) \\
\rowcolor{gray!15} \quad + both teachers & \textbf{54.4(0.4)}$^{**}$ & \textbf{72.2(0.3)}$^{**}$ & \textbf{5.1(1.4)}$^{**}$ & \textbf{58.3(1.0)}$^{**}$ \\
\addlinespace[2pt]
MoE-Retriever & 50.6(0.6) & 68.0(0.5) & 12.7(1.9) & 54.9(1.1) \\
\rowcolor{gray!15} \quad + both teachers & \textbf{52.1(0.8)}$^{*}$ & \textbf{70.1(0.3)}$^{**}$ & \textbf{3.8(0.9)}$^{**}$ & \textbf{56.1(0.9)} \\
\bottomrule
\end{tabularx}
\end{table}

\subsection{Calibration carries over to external cohorts}
\label{sec:aibl}
In discrimination, GRIP is on par with the strongest baselines (Table~\ref{tab:aibl}): it ranks third in balanced
accuracy and AUC on both cohorts, and it separates CN from AD best on AIBL and second
best on OASIS-3. Where GRIP stands out is calibration. With every method recalibrated on ADNI, GRIP and
XGBoost have the lowest ECE on AIBL, about half or less than that of every deep baseline, and on OASIS-3 GRIP is on par with the
lowest, that of XGBoost. M3Care and PRA-PoE, the
baselines with the highest AUC, have 1.7 to 3.4 times its calibration error. Even without recalibration, GRIP
is better calibrated on AIBL than every recalibrated deep baseline (Fig.~\ref{fig:reliability}c, d).

\begin{table}[t]
\centering
\caption{Ablation study of GRIP on the ADNI test set (mean (std) over 15 runs, \%). ECE of the trained models, before
any post-hoc calibration. W/o random draws: mean fills. $\gamma=1$: the second rollout as narrow as the first. W/o grading: no grading by the imputation teacher. W/o alignment: no neighborhood alignment. The
soft target is the diagnosis teacher and the graded rollouts the imputation teacher. Rank: the percentile rank, by cosine similarity, of a test visit's own real summary among other
participants' summaries when that modality is hidden and filled with the student's mean (mean over modalities and runs,
lower is better, chance 50, $^{\dagger}$at chance without fills). Best results in
\textbf{bold}, second best \underline{underlined}. $^{*}p<0.05$, $^{**}p<0.01$ versus GRIP in both tests of
Section~\ref{sec:protocol}.}
\label{tab:ablation}
\scriptsize
\setlength{\tabcolsep}{1.5pt}
\begin{tabularx}{\columnwidth}{@{}l*{5}{>{\centering\arraybackslash}X}@{}}
\toprule
 & \multicolumn{3}{c}{Seven combinations} & Real & Fills \\
\cmidrule(lr){2-4}\cmidrule(lr){5-5}\cmidrule(lr){6-6}
Model & ACC & AUC & ECE & ACC & Rank$\downarrow$ \\
\midrule
\rowcolor{gray!15} GRIP (full) & \textbf{55.4(0.5)} & \textbf{73.3(0.3)} & \underline{3.4(0.5)} & \textbf{59.6(0.7)} & 32.0 \\
\addlinespace[3pt]
\multicolumn{6}{@{}l}{\emph{Components of the graded rollouts}} \\
\quad w/o random draws & 54.6(0.6)$^{*}$ & 72.6(0.4)$^{**}$ & 6.6(1.7)$^{**}$ & 58.4(0.8)$^{*}$ & 32.8 \\
\quad w/o grading & 54.8(0.6)$^{*}$ & 72.6(0.4)$^{**}$ & 5.4(1.2)$^{**}$ & 59.0(0.9) & \textbf{31.5} \\
\quad w/o consistency & 54.7(0.4)$^{**}$ & 72.7(0.4)$^{**}$ & 4.1(0.9) & 59.1(0.9) & 32.0 \\
\quad $\gamma=1$ & 55.0(0.4) & 73.0(0.3)$^{*}$ & 3.8(0.8) & 59.3(0.9) & 32.0 \\
\quad w/o alignment & \textbf{55.4(0.5)} & 73.0(0.4)$^{**}$ & \textbf{3.2(0.4)} & \underline{59.4(0.7)} & 39.7 \\
\quad w/ teacher forcing & \textbf{55.4(0.5)} & \underline{73.2(0.3)} & 3.8(0.6)$^{**}$ & \textbf{59.6(0.6)} & 32.0 \\
\addlinespace[3pt]
\multicolumn{6}{@{}l}{\emph{Teachers}} \\
\quad w/o soft target & 54.8(0.5)$^{*}$ & 72.9(0.3) & 3.7(1.1) & 59.1(0.8) & \underline{31.9} \\
\quad w/o graded rollouts & \underline{55.1(0.5)} & 72.9(0.3)$^{*}$ & 4.5(0.7)$^{*}$ & 58.8(0.9) & 50.0$^{\dagger}$ \\
\quad w/o both teachers & 55.0(0.6) & 72.6(0.3)$^{**}$ & 5.3(1.4)$^{**}$ & 58.8(1.1) & 50.0$^{\dagger}$ \\
\bottomrule
\end{tabularx}
\end{table}

\subsection{The teachers improve other models}
\label{sec:plugin}
The two teachers change how a missing modality is filled and how each view is supervised, not GRIP's classifier,
so we add them to three recent methods with
different fusion designs: PRA-PoE, which fuses modalities in a product of experts, and UniMIX-AD and MoE-Retriever, which
route patients through mixtures of experts (Table~\ref{tab:plugin}). Each base keeps its encoders, architecture,
losses, optimizer, training views and model selection. For the graded rollouts, the base's own imputation module is
replaced by GRIP's rollout, whose fills are mapped back to the input features by the Stage-1 decoder, and for the soft
target, the distillation term is added to the base's loss.

With both teachers, the seven-combination accuracy and AUC of every base rise significantly, and its accuracy
under real missingness rises as well. The teachers also correct the poorly calibrated MoE-Retriever, whose ECE falls to about a third.

\subsection{What each part of GRIP contributes}
\label{sec:ablation}
\emph{Components.} Table~\ref{tab:ablation} removes or replaces one part of GRIP at a time. Drawing random fills
matters most: replacing the draws with their mean costs the most accuracy, as much AUC as any other component, and nearly
doubles the ECE, so training on several
plausible fills, rather than one estimate, improves both accuracy and calibration. Removing the teacher's grading
or the two-rollout consistency also lowers accuracy and AUC significantly, and removing the soft target lowers accuracy significantly
and AUC slightly: the fills help most when
they are plausible and the classifier is stable across them. The neighborhood alignment serves a different goal, tying
each fill to the individual patient: without it, the fill moves away from the patient's own data (Rank), while
accuracy is unchanged and AUC slightly lower. Compared with teacher forcing in every epoch, in which the context of the second fill is built from real
data, GRIP matches its accuracy and AUC and is significantly better calibrated. With one modality missing, the two grade alike, and with two missing, where they grade differently,
accuracy is unchanged and GRIP's ECE is lower, 3.5\% against 4.1\%. The two teachers
are complementary: the soft target adds more than three times as much accuracy
when the graded rollouts are present as when they are absent.

\section{Discussion and Conclusion}
\label{sec:discussion}

\emph{Limitations.} GRIP diagnoses each visit on its own. Many ADNI participants are seen several times, and their
history could inform both teachers, so a longitudinal extension is a natural next step. The imaging modality is a set
of regional MRI measurements rather than the images themselves, and extending the rollouts to image-level features
would test the idea where the missing modality is richest.  GRIP fills whole modalities. Because APOE is carried to every visit, most missing CSF values are replaced by the training mean, as for every
method, rather than drawn by the rollouts. Giving CSF its own
modality would let GRIP fill it, with rollouts of up to three steps. Longer rollouts, on cohorts with more modalities, remain to be tested. Ensembles and Monte Carlo dropout, which average several models or passes, could improve the calibration of
any method, GRIP included, at a multiple of the inference cost.

We introduced GRIP, in which two privileged teachers supervise one student for AD diagnosis under arbitrary
modality missingness: an imputation teacher grades each step of the student's own rollouts, and a diagnosis teacher
gives every combination a soft diagnosis. GRIP outperforms state-of-the-art baselines on ADNI, its teachers improve three
other models, and on two external cohorts it is the best or among the best calibrated, even after
every method is recalibrated. Since the teachers are needed only during
training, a single lightweight model serves every combination of tests a patient has received, and a low confidence
tells the clinician that the available tests leave the diagnosis open.

\section*{Acknowledgment}
Data were provided in part by OASIS-3: Principal Investigators: T. Benzinger, D. Marcus, J. Morris; NIH P50AG00561,
P30NS09857781, P01AG026276, P01AG003991, R01AG043434, UL1TR000448, R01EB009352. AV-45 doses were provided by Avid
Radiopharmaceuticals, a wholly owned subsidiary of Eli Lilly.

Data collection and
sharing for the Alzheimer's Disease Neuroimaging Initiative (ADNI) is funded by the National Institute on Aging
(National Institutes of Health Grant U19 AG024904).
The grantee organization is the Northern California Institute for Research and Education. In the past, ADNI has also
received funding from the National Institute of Biomedical Imaging and Bioengineering, the Canadian Institutes of
Health Research, and private sector contributions through the Foundation for the National Institutes of Health (FNIH)
including generous contributions from the following: AbbVie, Alzheimer's Association; Alzheimer's Drug Discovery
Foundation; Araclon Biotech; BioClinica, Inc.; Biogen; Bristol-Myers Squibb Company; CereSpir, Inc.; Cogstate; Eisai
Inc.; Elan Pharmaceuticals, Inc.; Eli Lilly and Company; EuroImmun; F. Hoffmann-La Roche Ltd and its affiliated
company Genentech, Inc.; Fujirebio; GE Healthcare; IXICO Ltd.; Janssen Alzheimer Immunotherapy Research \&
Development, LLC.; Johnson \& Johnson Pharmaceutical Research \& Development LLC.; Lumosity; Lundbeck; Merck \& Co.,
Inc.; Meso Scale Diagnostics, LLC.; NeuroRx Research; Neurotrack Technologies; Novartis Pharmaceuticals Corporation;
Pfizer Inc.; Piramal Imaging; Servier; Takeda Pharmaceutical Company; and Transition Therapeutics.

The MUSE regional volumes were obtained through the ADNI and AIBL databases.

During the preparation of this work, the authors used Claude (Anthropic) for language polishing. The authors
reviewed and edited the content as needed and take full responsibility for the content of the published article.

\bibliographystyle{IEEEtran}
\bibliography{refs}

\end{document}